# Classification of Resting-State fMRI using Evolutionary Algorithms: Towards a Brain Imaging Biomarker for Parkinson's Disease

Amir Dehsarvi[*]

Stephen L. Smith

Department of Electronic Engineering

University of York

YO10 5DD

---

[*] email address: ad953@york.ac.uk; Telephone No.: +44 (0)746 801 1161; ORCiD ID: 0000-0001-7116-9741

# Abstract

Accurate early diagnosis and monitoring of neurodegenerative conditions is essential for effective disease management and delivery of medication and treatment. This research develops automatic methods for detecting brain imaging preclinical biomarkers for Parkinson's disease (PD) by considering the novel application of evolutionary algorithms. A fundamental novel element of this work is the use of evolutionary algorithms to both map and predict the functional connectivity in patients using rs-fMRI data taken from the Parkinson's Progression Markers Initiative database (PPMI, a large-scale comprehensive PD dataset with age-matched healthy participants) to identify PD progression biomarkers. Specifically, Cartesian Genetic Programming was used to classify Dynamic Causal Modeling data as well as timeseries data. The findings were validated using two other commonly used classification methods (Artificial Neural Networks and Support Vector Machines) and by employing *k*-fold cross-validation. Across Dynamic Causal Modeling and timeseries analyses, findings revealed maximum accuracies of 75.21% for early stage (prodromal - occurring 5-20 years prior to motor symptoms) PD patients versus healthy controls, 85.87% for PD patients versus prodromal PD patients, and 92.09% for PD patients versus healthy controls. Prodromal PD patients were classified from healthy controls with high accuracy – this is notable and represents the key finding of this research since current methods of diagnosing prodromal PD have both low reliability and low accuracy. Furthermore, Cartesian Genetic Programming provided comparable performance accuracy relative to Artificial Neural Networks and Support Vector Machines. Evolutionary algorithms enable us to decode the classifier in terms of understanding the data inputs that are used, more easily than in Artificial Neural Networks and Support Vector Machines. Hence, these findings underscore the relevance of both Dynamic Causal Modeling analyses for classification and Cartesian Genetic Programming as a novel classification tool for brain imaging data with medical implications for disease diagnosis, particularly in early and asymptomatic stages.

PD Diagnosis

1. Introduction

In the UK, one in 500 people have Parkinson's disease (PD) and this is the 2nd most prevalent neurodegenerative disease globally, trailing only Alzheimer's disease [1]. There is poor differential diagnosis of neurodegenerative diseases with high rates of misdiagnosis and low test-retest reliability [2], [3]; indeed, PD has rates of misdiagnosis of 15-26% [4], [5], which is further exacerbated in the early stages of PD. Therefore, quick and non-invasive methods for diagnosis and monitoring with good accessibility are necessary. PD motor symptoms are apparent in the early phases of the disease, typically occurring following a loss of 60% of dopaminergic neurons [6], [7]. The premotor or prodromal PD phase is the period of time from the start of neurodegeneration and detection of clinical motor symptoms, between 5 and 20 years [8]. In the prodromal phase, individuals typically have non-motor symptoms (e.g., rapid eye movement, sleep behaviour disorder and olfactory dysfunction) [9]. Prodromal PD is typically misdiagnosed as another Parkinsonian disorder. Hence, a key step in ensuring accurate diagnosis involves early identification of prodromal PD patients, with key consequences for prompt patient treatment and disease management, which is examined in this research.

The UK spend £600 million yearly on economic costs associated with 100,000 PD patients (corresponding to £5993 per patient) [10]. Current expenditure is likely to be exacerbated by an ageing UK population with now almost 130,000 PD patients. Costs are increased with certain PD symptomology, ageing, and PD severity [11]–[13]. Indirect financial costs associated with PD include loss of earnings via early retirement of the PD patient or a family member who cares for them. Until a cure for PD is developed, early diagnosis and treatment are essential in improving quality of life for PD patients and their family, which will help to reduce any financial burdens linked to PD on both the patients' family and the NHS.

This research examines a PD biomarker to facilitate prompt diagnosis by developing automatic methods of diagnosing early stage PD via examining the classification of resting state fMRI using evolutionary algorithms (EAs).

PD Diagnosis

## 1.1. Diagnosis

Currently, the Movement Disorder Society Task Force for Rating Scales for Parkinson's Disease (MDS-UPDRS[2]) [14] is widely used to diagnose PD and includes measures of motor and cognitive deterioration. Nevertheless, these diagnostic tests have limited objectivity due in part to insufficient medical training regarding delivering the MDS-UPDRS [15]. Indeed, rates of accurate diagnoses are particularly low in both primary care by GPs [2] and secondary care by specialists [3], resulting in patients obtaining inadequate treatment. A challenge involves differentiating PD from alternative diseases that present similar symptomologies; a difficulty that is compounded in the early phases of PD. PD has rates of misdiagnosis of 25% as it is often mistakenly labelled as another neurodegenerative disease, for example, progressive supranuclear palsy [3], [16]. Research reveals that 10% of PD cases are misdiagnosed as atypical Parkinsonism or a Parkinson's plus syndrome [17]. Further, post mortem research has revealed that up to 15-26% of PD cases were misdiagnosed by general neurologists [4], [5], with only 8-15% of misdiagnosis when diagnosed by expert movement disorder clinics [3], [18], [19]. Yet, due to limited financial and time-related resources, it is not feasible for movement disorder experts to diagnose each case of potential neurodegenerative diseases. This research identifies an imaging biomarker for PD using functional brain imaging (fMRI), an ideal tool that can increase diagnostic accuracy dramatically given that it is both non-invasive and not reliant on diagnostic tests that may involve subjective evaluations. Objective evaluations can be challenging, specifically given the complexity of certain symptoms, such as bradykinesia. Specialised equipment, for instance, learning algorithms may be better equipped to detect early stages of cognitive and motor decline, relative to medics.

## 1.2. Learning Algorithms

A component of assessing motor decline involves a finger-tapping task in which patients tap their thumb and forefinger numerous times (as widely and as fast as possible) whilst a medic evaluates performance. In this task, inter- and intra-rater reliability is typically low, attributed to the limited capacity of medics to evaluate performance solely via observation [20]. A new approach involves capturing patient movements via computational techniques and using evolutionary algorithms (EAs) to assess performance. Research has revealed that EAs applied to performance data from a finger tapping task alone differentiated PD patients from age-matched controls with an accuracy of ~95% [21]. As such, data from traditional motor exams combined with new computational techniques are a valid and relatively novel approach in diagnosing PD.

---

[2] The Movement Disorder Society Task Force for Rating Scales for Parkinson's Disease prepared a critique of the Unified Parkinson's Disease Rating Scale. Advantages of this scale are that it is frequently used in western countries, and that is can be used across the clinical spectrum of PD since it addresses motor symptoms. In addition, this scale has good clinometric properties, including reliability and validity. Disadvantages include ambiguities in the instructions, several metric limitations, and zero screening questions on various significant non-motor elements of PD.

Hence, this research examines a key question: Can early stage PD be diagnosed using EAs on rs-fMRI (resting state fMRI) data? Early diagnosis of PD is fundamental in providing patients with palliative care during the early phases before motor symptoms are present, enabling effective disease management and maintaining patient quality of life. Moreover, once a neuroprotective drug to treat PD is developed, the early diagnosis of PD would have even greater clinical implications [22].

Research has explored the classification of PD patients using learning algorithms, including EAs. EAs offer a novel approach to disease classification. EAs are optimising algorithms based on Darwinian evolutionary theory. Cartesian Genetic Programming (CGP) is a subtype of EAs that as a norm evolves directed acyclic computational structures of nodes. Recurrent CGP (RCGP) is an extension of CGP, which enables cyclic or feedback connections. CGP and RCGP have never been applied to neuroimaging data, including rs-fMRI, hence, their applicability to this data is examined in this research.

A key benefit of using EAs alongside an expressive dynamical representation is the ability to explore a wide area of classifiers. In addition, since these classifiers do not rely on expert knowledge, they can identify trends that might not be detected by experts and contribute to furthering expert knowledge. For instance, evolved classifiers and their distributions have provided the following scientific contributions: the differential effect of dominance on diagnostic accuracy, the over-representation of certain trends of acceleration in the movements of PD patients, and amplitude and frequency blends with diagnostic power. Regardless of how efficient these classifiers are, a limitation of this method is the lack of knowledge underpinning how these algorithms function, rendering them often unfathomable to experts. Hence, these classifiers are a valuable tool in guiding and/or supporting a medical diagnosis, yet, an automated diagnosis cannot be approved unless the clinician is confident regarding the biological underpinnings of the diagnosis.

This research explores an automatic and non-invasive method of confirming the diagnosis of PD, specifically, examining the classification of participants diagnosed with PD, prodromal PD participants, and healthy age-matched controls using rs-fMRI data. This research involves an analysis of rs-fMRI data taken from the Parkinson's Progression Markers Initiative database (PPMI; http://www.ppmi-info.org/data). Much research has been conducted to identify prodromal, motor stage and other biomarkers, with limited success. Possible reasons for this limited success involve the participant sample characteristics (PD research typically focuses on patients with a confirmed diagnosis and in the late stages of the disease), diagnostic criteria, and poor storage and collection of data [23], [24]. PPMI focuses on eliminating many of these key limitations by aiming to recruit equal numbers of early stage (pre-medication) PD patients and healthy age-matched controls [25]. The PPMI [25] is a landmark, large-scale, comprehensive, observational, international, and multi-centre study that recruits *de novo* (early-untreated) PD patients, prodromal PD patients, and age-matched healthy participants (among other participant groups) to identify PD progression biomarkers. PPMI contains 20 centres across the EU and the USA and these follow standardised procedures for repeated bio-sampling (blood, CSF, urine), clinical assessments, and imaging as well as rigorous standards for data storage and analysis.

### 1.3. rs-fMRI Clinical Data

The current research applies EAs, specifically CGP and RCGP, for the classification of rs-fMRI in PD using Dynamic Causal Modeling (DCM) [26]–[29] and timeseries analyses. DCM is a powerful tool that explores effective connectivity (the causal effect of one neuronal system on another) using nonlinear designs to identify a reasonable generative model of measured neural activity (electromagnetic measurements or hemodynamic fMRI measurements). DCM contains information regarding how neuronal activity results in the measured responses, which allows estimation of the effective connectivity. The current research examines DCM for rs-fMRI, in which deterministic inputs are activating/causing changes in the stimulation of different brain regions. This occurs via a dynamic input–state–output model of several inputs and outputs. The inputs relate to standard stimulus functions linked to the experimental manipulations. In this research, the timeseries values and DCM values from the rs-fMRI data are subjected to supervised classification and the findings are validated with two other commonly used classification methods (Artificial Neural Networks, ANN, and Support Vector Machines, SVM) as well as employing *k*-fold cross-validation (CV).

A key aim of this research is to identify the applicability of CGP and RCGP classification for both timeseries and DCM analyses regarding the analysis of PD data. CGP and RCGP have not previously been used in the classification of brain imaging data. This study examines an additional novel question: is DCM analysis useful for classification for PD data? Previous research has not explored the applicability of DCM values in classification and, to date, little research has applied DCM to PD data [30]–[35]. Hence, by doing both, this research develops automatic procedures for identifying PD brain imaging preclinical biomarkers, which can be used for aiding/confirming early PD diagnosis.

## 1.4. Class-Imbalanced Samples

Furthermore, a typical difficulty when conducting medical research involves recruiting equal sample sizes of patients and healthy controls, often resulting in class-imbalanced data (e.g., unequal groups of controls versus patients). The PPMI database is heavily class-imbalanced, with many more PD patients relative to prodromal PD and control participants. Learning algorithms typically assume approximately equal class distributions. Hence, these algorithms may function with low accuracy when faced with significantly imbalanced datasets, labelled between-class imbalance. As such, highly accurate classifiers that function with between-class imbalanced data to identify a minority class (e.g., PD patients) are necessary. In these cases, the overall accuracy or error rate may not be sufficient and other metrics (e.g., receiver operating characteristics curves, precision-recall curves, and cost curves) may better represent the performance of algorithms with imbalanced data. Small sample sizes with significantly imbalanced class distributions are particularly common in clinical research due to challenges in recruiting patient samples and constrict learning due to two limitations. Firstly, reduced sample sizes result in problems linked to absolute rarity and within-class imbalances. Secondly, algorithms frequently do not generalise inductive rules across the sample. Algorithm performance is restricted by the limitations implicit in generating conjunctions across many features with reduced samples, which can result in overfitting (which occurs when the rules produced are overly precise).

Indeed, the class-imbalance problem is a relative problem that depends on (1) the degree of class-imbalance; (2) the complexity of the concept represented by the data; (3) the overall size of the training set; and (4) the classifier involved. More specifically, the higher the degree of class-imbalance, the higher the complexity of the concept. The smaller the total size of the training set, the greater the effect of class imbalances in classifiers sensitive to the problem [36].

Research has revealed that the performance of certain classifiers trained on specific imbalanced data can be similar to the performance of these same classifiers trained on the same data that has been modified to have approximately equal class distributions [36], [37]. Yet, for most imbalanced data, solutions for learning predominantly focus on modifying the data sample to achieve a sample that has balanced class distributions, which enhances the overall classification accuracy relative to the original imbalanced sample [38]–[40]. This research applies ADASYN [41] to generate class balanced data for classification. ADASYN systematically and adaptively generates varying numbers of synthetic data based on minority and majority class distributions to create class balanced data [41], focusing on generating synthetic data close to the inter-class boundary.

Hence, this research explored the applicability of classification methods to class-imbalanced data, with key implications for the transferability of medical research based on limited and imbalanced sample sizes.

### 1.5. Research Overview

The work presented develops automatic methods for identifying PD brain imaging preclinical biomarkers, which can aid clinical diagnosis, monitoring and investigation of PD. Currently prodromal PD diagnosis (before motor symptoms are apparent) is in its infancy with typically low accuracy rates and high levels of misdiagnosis with other Parkinsonian conditions. Hence, prodromal PD diagnosis is highly relevant, ensuring access to early treatment for patients before motor symptoms appear and providing overall better disease management, thus, increasing patient quality of life. Specifically, this research examines the following question: Can early stage PD be diagnosed using EAs on rs-fMRI data?

This research uses an exploratory and data-driven approach to develop novel clinical monitoring tools and applies these to the diagnosis of participants with PD, early stage PD, and healthy age-matched controls. Developing a tool that can differentiate between the various stages of disease severity has therapeutic consequences in terms of tailoring medication dosage and monitoring medication in accordance with symptoms exhibited and overall PD stage. The research presented analyses open data taken from the PPMI, a longitudinal study where participants underwent a comprehensive longitudinal follow-up schedule of clinical, imaging and bio-specimen assessments.

This research develops automatic procedures for identifying PD brain imaging preclinical biomarkers, which enhances the confidence of methods involved in early PD diagnosis. A core research aim is to identify the applicability of CGP and RCGP classification for both timeseries and DCM analyses. The timeseries values and DCM values from the rs-fMRI data are subjected to supervised classification and the findings are validated with two other commonly used classification methods (ANN and SVM). EAs, such as CGP and RCGP, have not previously been applied to brain imaging data. A crucial advantage of EAs, specifically CGP, is that they offer a *white box solution* providing more information on the inputs used and better understanding of the final solution obtained in classification, relative to ANN and SVM. Moreover, research on the classification of rs-fMRI data has typically used statistical-based classifiers (e.g., independent components analysis and multivariate pattern analysis [42]–[46]; for an example of independent components analysis in rs-fMRI for PD data, see [47]). This research examines an additional novel question: is DCM analysis useful for classification? Previous research has not examined the applicability of DCM values in classification and little research has applied DCM to PD data [30]–[35].

A common limitation with medical data involves recruiting low numbers of patients, which can result in class-imbalanced data (e.g., high numbers of controls versus patients). This research examines the applicability of classification methods to two datasets with heavily class-imbalanced data, which mimics the conditions prolific in medical research, enabling the research findings to be more easily generalised to clinical settings.

## 2. Method

### 2.1. Participants

PPMI is a longitudinal study where participants underwent a comprehensive longitudinal follow-up schedule of clinical, imaging and biospecimen assessments. There were eight healthy controls (all male, mean age = 68, $SD$ = 3.16), 18 prodromal PD patients (13 male, mean age = 68, $SD$ = 4.03), and 102 early PD patients (71 male, mean age = 63, $SD$ = 7.83). The overall age range was 50-75 years. 3 Tesla rs-fMRI, dopamine transporter (DAT) imaging, and MRI scans were acquired for all participants.

PD participants were recruited at disease threshold (diagnosis within two years and untreated for PD) and were required to have an asymmetric resting tremor or asymmetric bradykinesia or two of bradykinesia (resting tremor and rigidity). DAT deficit was acquired for PD participants. Healthy participants had no significant neurologic dysfunction, no first degree family member with PD, and they obtained a MoCA > 26. The study was approved by the institutional review board of all participating sites. Written informed consent was obtained from all participants.

### 2.2. rs-fMRI Acquisition

A standardised MRI protocol included acquisition of whole-brain structural and functional scans on 3 Tesla Siemens Trio Tim MR system (for more information see http://www.ppmi-info.org/). 3D T1 structural images were acquired in a sagittal orientation using a MPRAGE GRAPPA protocol with Repetition Time (TR) 2300 ms, Echo Time (TE) 2.98 ms, Field of View (FoV) 256 mm, Flip Angle (FA) 9° and 1 $mm^3$ isotropic voxel. For each participant, 212 BOLD echo-planar rs-fMRI images (40 slices each, ascending direction) were acquired during an 8 min, 29 s scanning session (acquisition parameters: TR = 2400 ms, TE = 25 ms, FoV = 222 mm, FA = 80° and 3.3 $mm^3$ isotropic voxels). Participants were instructed to rest quietly, keeping their eyes open, and were asked not to fall asleep.

## 2.3. Imaging Data Analysis

### 2.3.1. Preprocessing

The imaging data analyses were done using the CONN (version 17.c) [48] and SPM12 (version 6906 - Wellcome Department of Imaging Neuroscience, London, UK) [49] software packages in MATLAB. Preprocessing included DICOM to 3D NIFTI conversion and reduction of the spatial distortion using Field Map toolbox in SPM12 [49]. Anatomical data was segmented; both anatomical and functional data were normalised. All the functional images were motion corrected and coregistered to participants' own high-resolution anatomical image. The participants' anatomical images were normalised to the standard T1 template in the Montreal Neurological Institute (MNI) space as provided by SPM12. Then the normalisation parameters of each participant were applied to the functional images to normalise all the functional images into the MNI space. The EPI data were unwarped (using field-map images) to compensate for the magnetic field inhomogeneities, realigned to correct for motion, and slice-time corrected to the middle slice. The normalisation parameters from the T1 stream were then applied to warp functional images into MNI space. All the functional images were spatially smoothed using a Gaussian kernel with 8 mm [50] FWHM to account for inter-participant variability while maintaining a relatively high spatial resolution. Linear and quadratic detrending of the fMRI signal was applied, which involved covarying out white matter (WM) and Cerebrospinal fluid (CSF) signal. WM and CSF signals were predicted for each volume from the mean value of WM and CSF masks derived by thresholding SPM's tissue probability maps at 0.5. The resting data were bandpass filtered (0.008–0.1 Hz). The main analysis used spectral DCM as per SPM12.

### 2.3.2. Processing

#### 2.3.2.1. Timeseries

Functional connectivity in the Default Mode Network (DMN) is well studied. Hence, this research took as regions of interest (nodes) the most commonly reported four major parts of DMN, as shown in Figure 1: medial prefrontal cortex (mPFC, centred at 3, 54, −2), posterior cingulate cortex (PCC, centred at 0, −52, 26), left inferior parietal cortex (LIPC, centred at −50, − 63, 32), and right inferior parietal cortex (RIPC, centred at 48, − 69, 35). For each

PD Diagnosis

participant, the volumes of interest were defined as spheres centred at those coordinates mentioned above with an 8 mm [50] radius and with a mask threshold of 0.5. The first eigenvectors were extracted after removing the effect of head motion and low frequency drift. This vector is stored for each region as timeseries.

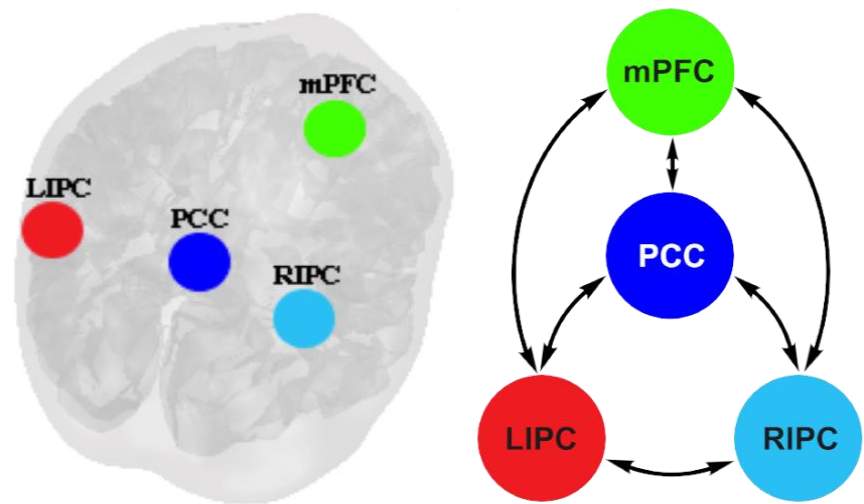

Figure 1 - The four DMN regions of interest used in this research

### 2.3.2.2. Dynamic Causal Modeling (DCM)

The spectral DCM analyses [51] were conducted using the DCM12 implemented in SPM12. The regions of interest of DCM analyses were defined according to the peak of the DMN independent component maps, as presented in Figure 1. The main purpose of the current DCM analysis was to investigate the endogenous/intrinsic effective connectivity and to examine the causal interactions across these regions. The modelled low frequency fluctuations were set as driving inputs to all four nodes, and different models were defined by considering a full connection for all nodes. Expected posterior model probabilities and exceedance probabilities were computed. The intrinsic connectivity parameters (16 values that were stored in DCM.Ep.A matrix, all parameters of intrinsic/effective connectivity [28]) from each participant were subjected to classification using CGP.

## 2.4. Cartesian Genetic Programming (CGP)

### 2.4.1. Classification

There was one output for the classifier (class 1 for one group, class 0 for the other one, e.g., class 1 for PD patients versus class 0 for healthy controls). To have equal class representation, data from each class was divided randomly into subsets of 70% (training), 15% (validation), and 15% (test). The geometry of the programs in the population (chromosomes) has fifty nodes with a function set of four mathematical operations (+, -, ×, ÷), multiple inputs (according to the datasets), and one output (either class 1 or class 0 for each binary combination of participant groups). At each generation, the fittest chromosome is selected and the next generation is formed with its mutated versions (mutation rate = 0.1). Evolution stops when 15000 iterations are reached. To obtain statistical significance, the classification was done in 10 runs for each combination of inputs and the accuracy was averaged over the runs. The results (the winning chromosome, the networks, and the accuracy values) were stored for each run individually.

#### 2.4.1.1. Classification of Timeseries and DCM

rs-fMRI is a widespread tool for exploring the functionality of the brain, using volume timeseries data. These scans contain abundant data; hence, obtaining relevant and useful data from raw scans (i.e., high dimensional datasets) can be difficult. Machine learning algorithms provide various tools that create datasets with less dimensions and more useful data, although challenges persist regarding how to select relevant data and how to maintain the interpretability of this data. This can result in losing important properties of the raw data, although dealing with such a large number of features can be computationally expensive and very time consuming.

In the current experiment, the RCGP algorithm is used to classify the features that appeared across time in participant scans. The number of timeseries values is 210, i.e., for each of the four regions in the DMN, there is a vector of 210 values. Analysing/classifying the timeseries values was conducted in three different ways in terms of inputs to the classifier:

PD Diagnosis

1. The timeseries values for each region separately were used as inputs to the classifier, to classify with 210 features per region (relating to DMN regions: LIPC, RIPC, PCC, and mPFC) and per participant.
2. The timeseries values together in four columns (one per region) were used as inputs to the classifier to classify the data with four columns (corresponding to the four DMN regions) of 210 features per participant.
3. The timeseries values inserted together in one column were used as inputs to the classifier to classify the data with 840 features in one vector for each participant. The order of inputting the timeseries values for each DMN region to form the final vector was consistent between participants.

Classification was completed in 10 runs for each combination of inputs and the accuracy was averaged over the runs. The same inputs were used to classify the data using ANN and SVM in MATLAB for comparison/validation. RCGP was used with 10% probability for the recurrent connections. A complete pipeline of the preprocessing and processing of the data is presented in Figure 2.

For this research, a new open source cross platform CGP library (version 2.4) [52] was used since it is able to evolve symbolic expressions, Boolean logic circuits, and ANN, and it can be extended to different areas. The CGP library enables the control of evolutionary parameters and the application of custom evolutionary stages.

PD Diagnosis

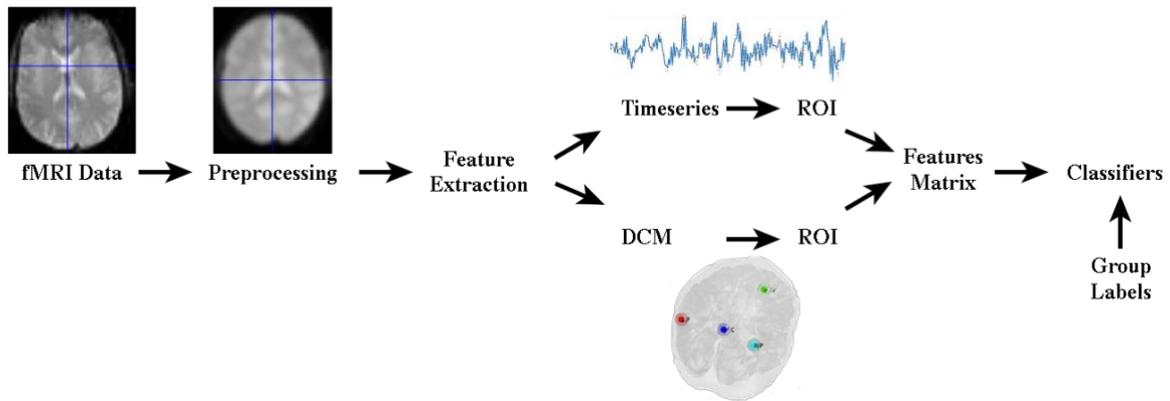

Figure 2 - Data analysis pipeline used in this research

### 2.4.1.2. Classification of DCM

The classification was run in the CGP Library with 16 inputs (all the DCM values sorted by region and presented as only one vector per participant). To facilitate comparison, the classification with data from the same participants was run using ANN and SVM, both run in MATLAB.

### 2.5. Adaptive Synthetic Sampling (ADASYN)

As previously mentioned, the control group contained eight participants, the prodromal PD group contained 18 patients, and the PD group contained 102 patients. Therefore, the data was highly imbalanced. Adaptive synthetic sampling was used to make the data balanced for training the classifier. After the process, the minor group for each combination (in the training set) had a higher number of participants, which made the data balanced for CV. The participant numbers in the validation and test sets were kept the same.

### 2.6. *k*-Fold Cross-Validation

*10*-fold CV was used to evaluate the classification accuracy using an unbiased estimate of the generalisation accuracy [53] excepting that for both PD participants and prodromal PD participants relative to control participants, CV was completed using *9* folds (due to the small number of samples in the control group). CV is beneficial as it allows the generation of independent test sets with enhanced reliability. With *10*-fold CV, typically one (of *10*) subset is the test set and remaining nine subsets are training sets. These sets are then rotated so that each set is used to test the data once. One repetition of the *10*-fold CV does not produce sufficient classification accuracies for comparison, therefore, *10*-fold CV is repeated 10 independent times and the mean accuracy over all *10* trials is calculated.

Since the main classification methodology in this research involved dividing the data into three different subsets (training, validation, and test), the data was divided into *10* subsets and, each time, one of the *10* subsets were used as the test set, another one for validation, and the remaining eight as a training set. The data was divided using stratified random sampling enabling the sample proportion in each data subset to be the same as that in the original data (i.e., equal class distribution in the subsets as per the original data). Hence, the data was split for each class initially and then the classes were mixed to form the completed set. This was done for each combination of inputs to both CGP (for DCM values) and RCGP (for timeseries values). In this study, the data for each combination of inputs was divided into three parts of 80% (training), 10% (validation), and 10% (test).

## 3. Results

This study examined classification of 102 PD participants, 18 prodromal PD participants, and eight healthy age-matched controls. The analysis (classification) focused on organising features to be used as inputs to the classifier in CGP and also in RCGP and was implemented using the CGP Library. To validate these findings, the analysis/classification was additionally completed using ANN and SVM in MATLAB.

### 3.1. Classification of Timeseries

Initially, the timeseries values for each region were used as inputs to the classifier individually. Therefore, the data was classified with 210 features from each region per participant. The same procedure was repeated separately for each DMN region (PCC, mPFC, RIPC, and LIPC). Then, the timeseries values were used as inputs to the classifier to classify the data with four columns (relating to the four DMN regions) of 210 features per participant. Finally, the timeseries values together in one column, were used as inputs to the classifier to classify the data with 840 features in one vector for each participant. The results after 10 runs for each combination were

PD Diagnosis

averaged and are presented for each category in Table 1, Table 2, and Table 3. The classification was also completed using ANN and SVM. For SVM, only the training and test sets were considered for classification.

Table 1 - Classification results for the timeseries values (PD vs. controls)

| | | Training % (*SD*) | Validation % (*SD*) | Test % (*SD*) |
|---|---|---|---|---|
| Classification results for each DMN region | | | | |
| PCC | RCGP | 92.06 (2.67) | 91.60 (2.91) | 91.95 (2.66) |
| | ANN | 92.73 (0.07) | 92.72 (0.32) | 92.74 (0.31) |
| | SVM | 92.73 (0.00) | NA | 92.73 (0.00) |
| mPFC | RCGP | 92.03 (2.68) | 91.73 (3.02) | 92.01 (2.88) |
| | ANN | 92.69 (0.06) | 92.85 (0.41) | 92.76 (0.40) |
| | SVM | 92.73 (0.00) | NA | 92.73 (0.00) |
| RIPC | RCGP | 91.13 (3.97) | 92.41 (3.64) | 91.46 (5.15) |
| | ANN | 92.68 (0.11) | 92.85 (0.50) | 92.90 (0.39) |
| | SVM | 92.73 (0.00) | NA | 92.73 (0.00) |
| LIPC | RCGP | 91.66 (2.55) | 91.75 (2.76) | 91.68 (3.46) |
| | ANN | 92.69 (0.10) | 92.89 (0.39) | 92.61 (0.38) |
| | SVM | 92.73 (0.00) | NA | 92.73 (0.00) |
| Classification results for all the DMN regions (4 inputs) | | | | |
| | RCGP | 91.89 (2.77) | 92.03 (3.34) | 91.79 (3.45) |
| | ANN | 92.66 (0.13) | 92.96 (0.45) | 92.74 (0.46) |
| | SVM | 92.78 (0.01) | NA | 92.73 (0.00) |
| Classification results for all the DMN regions (1 input) | | | | |
| | RCGP | 89.95 (8.53) | 91.84 (2.77) | 92.09 (2.68) |
| | ANN | 92.75 (0.08) | 92.74 (0.25) | 92.59 (0.07) |
| | SVM | 92.73 (0.00) | NA | 92.73 (0.00) |

Table 2 - Classification results for the timeseries values (PD vs. prodromal PD)

| | | Training % (*SD*) | Validation % (*SD*) | Test % (*SD*) |
|---|---|---|---|---|
| Classification results for each DMN region | | | | |
| PCC | RCGP | 85.64 (2.03) | 85.94 (2.23) | 85.87 (2.24) |
| | ANN | 85.02 (0.18) | 84.80 (0.82) | 85.02 (0.48) |
| | SVM | 85.00 (0.00) | NA | 85.00 (0.00) |
| mPFC | RCGP | 85.10 (1.67) | 85.97 (2.19) | 85.49 (2.35) |
| | ANN | 85.00 (0.13) | 84.66 (0.33) | 85.25 (0.69) |
| | SVM | 85.00 (0.00) | NA | 85.00 (0.00) |
| RIPC | RCGP | 84.93 (1.22) | 87.54 (1.12) | 80.64 (11.25) |
| | ANN | 84.97 (0.07) | 85.10 (0.28) | 84.93 (0.21) |
| | SVM | 85.00 (0.00) | NA | 85.00 (0.00) |
| LIPC | RCGP | 75.00 (22.97) | 85.85 (2.36) | 85.56 (3.15) |
| | ANN | 84.99 (0.15) | 84.91 (0.62) | 85.05 (0.47) |
| | SVM | 85.00 (0.00) | NA | 85.00 (0.00) |
| Classification results for all the DMN regions (4 inputs) | | | | |
| | RCGP | 84.69 (2.87) | 87.54 (4.01) | 85.02 (3.51) |
| | ANN | 85.06 (0.15) | 84.74 (0.64) | 84.95 (0.46) |
| | SVM | 85.06 (0.02) | NA | 85.00 (0.01) |
| Classification results for all the DMN regions (1 input) | | | | |
| | RCGP | 85.47 (1.86) | 86.27 (2.78) | 85.71 (2.43) |
| | ANN | 84.99 (0.03) | 85.12 (0.34) | 84.90 (0.26) |
| | SVM | 85.00 (0.00) | NA | 85.00 (0.00) |

PD DiagnosisTable 3 - Classification results for the timeseries values (prodromal PD vs. controls)

| | | Training % (*SD*) | Validation % (*SD*) | Test % (*SD*) |
|---|---|---|---|---|
| Classification results for each DMN region | | | | |
| PCC | RCGP | 69.46 (14.76) | 78.74 (7.12) | 62.41 (18.16) |
| | ANN | 69.05 (0.42) | 69.16 (2.09) | 68.53 (1.44) |
| | SVM | 69.23 (0.01) | NA | 69.22 (0.02) |
| mPFC | RCGP | 66.82 (15.66) | 76.32 (8.82) | 66.49 (17.38) |
| | ANN | 69.72 (0.79) | 69.53 (1.78) | 69.32 (1.75) |
| | SVM | 69.87 (0.13) | NA | 69.73 (0.26) |
| RIPC | RCGP | 55.93 (25.99) | 83.26 (10.26) | 55.15 (21.75) |
| | ANN | 69.42 (0.32) | 69.53 (1.62) | 70.11 (2.02) |
| | SVM | 69.70 (0.14) | NA | 69.23 (0.25) |
| LIPC | RCGP | 67.02 (14.68) | 78.66 (16.73) | 70.62 (13.40) |
| | ANN | 69.18 (0.50) | 69.17 (1.70) | 68.92 (1.56) |
| | SVM | 69.23 (0.00) | NA | 69.23 (0.00) |
| Classification results for all the DMN regions (4 inputs) | | | | |
| | RCGP | 62.45 (12.09) | 72.86 (10.24) | 62.46 (22.83) |
| | ANN | 69.35 (0.74) | 68.96 (1.83) | 69.75 (1.16) |
| | SVM | 74.32 (0.27) | NA | 70.76 (0.52) |
| Classification results for all the DMN regions (1 input) | | | | |
| | RCGP | 64.15 (15.10) | 70.04 (12.18) | 54.64 (30.34) |
| | ANN | 69.27 (0.15) | 68.77 (0.74) | 69.35 (0.66) |
| | SVM | 69.25 (0.01) | NA | 69.21 (0.06) |

As depicted in Table 1, findings revealed that PD patients were successfully classified from healthy controls with a maximum of 92.09% accuracy using RCGP (minimum accuracy: 91.46%). The results from the other two classification techniques (ANN and SVM) validated this finding as they were very similar: between 92.59% and 92.90% for ANN and SVM in all the different combinations of inputs.

PD Diagnosis

Table 2 illustrates that PD patients were successfully classified from prodromal PD patients with a maximum accuracy of 85.87% using RCGP (minimum accuracy: 80.64%). The results from the other two classification techniques (ANN and SVM) validated this finding as they were very similar: between 84.90% and 85.25% for ANN and SVM in all the different combinations of inputs.

The results revealed, as represented in Table 3, that prodromal PD patients were successfully classified from healthy controls with a maximum accuracy of 70.62% using RCGP (minimum accuracy: 54.64%). The results from the other two classification techniques (ANN and SVM) validated this finding as they were very similar: between 68.53% and 70.76% for ANN and SVM in all the different combinations of inputs.

Unlike for the DCM analyses, mixed ANOVAs were not conducted to evaluate the correspondence between participant group and timeseries features given that there were 840 features per participant, which would not be interpretable.

### 3.2. Classification of Dynamic Causal Modeling (DCM)

Classification using CGP implemented in CGP Library was executed with 16 inputs (all the DCM values sorted by region and presented as only one vector per participant) and 1 output (class 1 for one group, class 0 for another one). The results were then averaged over 10 runs and are presented for each category in Table 4. The classification was also done using ANN and SVM. For SVM, only the training and test sets were considered for classification.

Table 4 - Classification results for DCM values

|  | Training % (*SD*) | Validation % (*SD*) | Test % (*SD*) |
|---|---|---|---|
| PD vs. controls | | | |
| CGP | 91.00 (5.56) | 93.23 (3.83) | 90.87 (4.41) |
| ANN | 93.42 (1.97) | 93.52 (7.57) | 91.18 (8.86) |
| SVM | 92.86 (0.68) | NA | 92.86 (0.68) |
| PD vs. prodromal PD | | | |
| CGP | 80.01 (7.79) | 90.82 (3.85) | 79.12 (12.36) |
| ANN | 86.06 (2.47) | 83.32 (8.29) | 83.32 (7.39) |
| SVM | 85.36 (0.58) | NA | 85.36 (0.58) |
| Prodromal PD vs. controls | | | |
| CGP | 74.11 (27.42) | 90.11 (12.31) | 75.21 (23.01) |

PD Diagnosis

| | | | |
|---|---|---|---|
| ANN | 78.34 (16.44) | 70.00 (22.97) | 65.00 (26.87) |
| SVM | 68.42 (0.00) | NA | 68.42 (0.00) |

As illustrated in Table 4, findings revealed that PD patients were successfully classified from healthy controls with 90.87% accuracy using CGP. The results from the other two classification techniques (ANN and SVM) validated this finding as they were very similar: 91.18% for ANN and 92.86% for SVM. The results also revealed that PD patients were classified from prodromal PD patients with 79.12% accuracy using CGP. The results from ANN and SVM again validate these findings with 83.32% and 85.36% accuracy rates, respectively. Finally, the results indicated that prodromal PD patients were classified from healthy controls with 75.21% accuracy using CGP. The results from ANN and SVM also validate these findings with 65.00% and 68.42% accuracy rates, respectively.

Examples of the CGP classification trees/graphs can be seen in Figure 3 and Figure 4 for PD versus controls, Figure 5 and Figure 6 for PD versus prodromal PD, and Figure 7 and Figure 8 for prodromal PD versus controls. This represents a fundamental benefit of CGP, generating a *white box solution* that enhances interpretability of the classification network (not always possible with ANN and SVM classification methods). Despite the complexity inherent in these networks, they can provide crucial relevant information. For instance, in Figure 3 and Figure 4, approximately half of the inputs have been used to arrive at the final classification. These two networks are rather simpler than those represented in Figure 5, Figure 6, Figure 7, and Figure 8.

PD Diagnosis

Figure 3 - CGP classification tree for the classification of PD vs. controls; example 1

PD Diagnosis

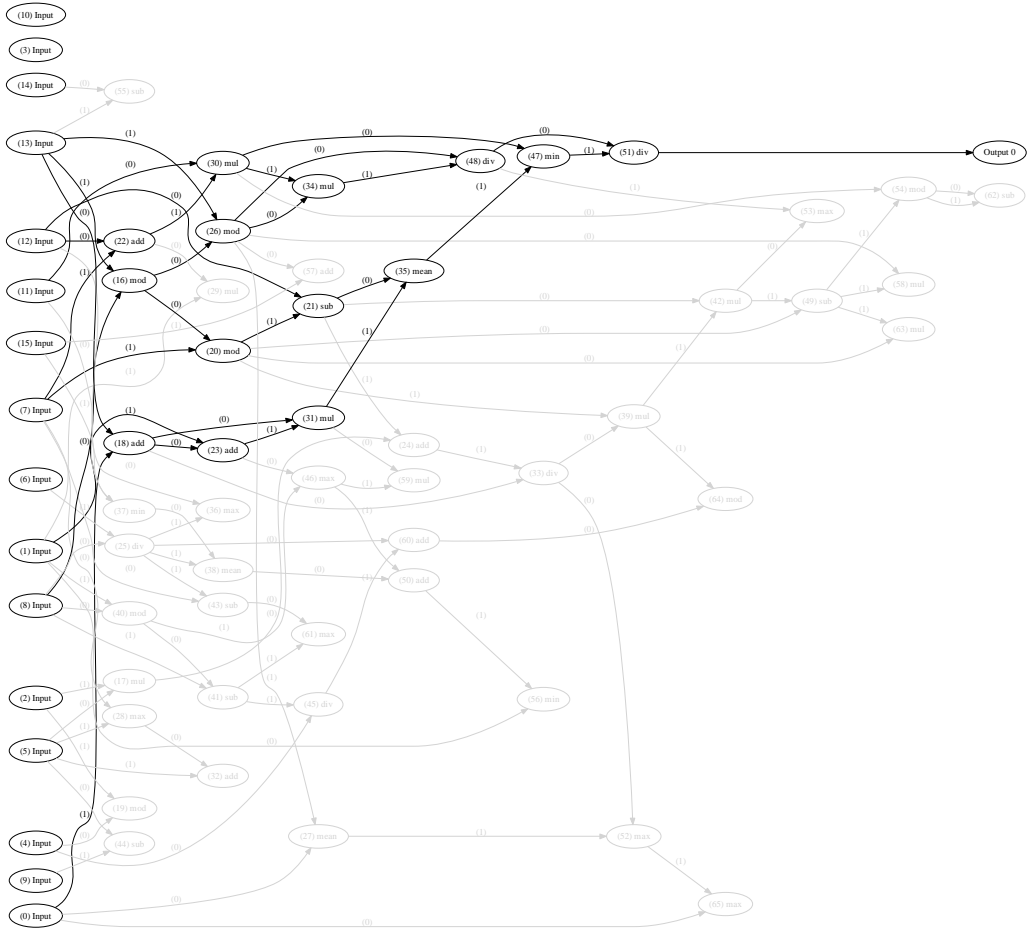

Figure 4 - CGP classification tree for the classification of PD vs. controls; example 2

PD Diagnosis

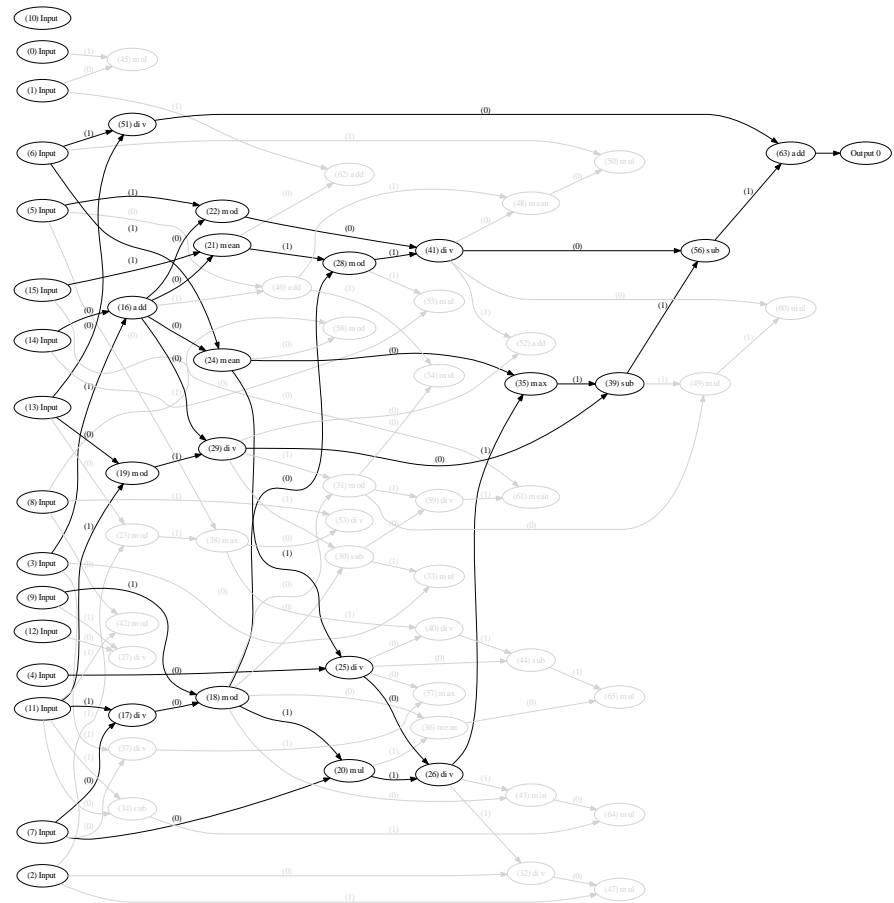

Figure 5 - CGP classification tree for the classification of PD vs. prodromal PD; example 1

PD Diagnosis

Figure 6 - CGP classification tree for the classification of PD vs. prodromal PD; example 2

PD Diagnosis

Figure 7 - CGP classification tree for the classification of prodromal PD vs. controls; example 1

PD Diagnosis

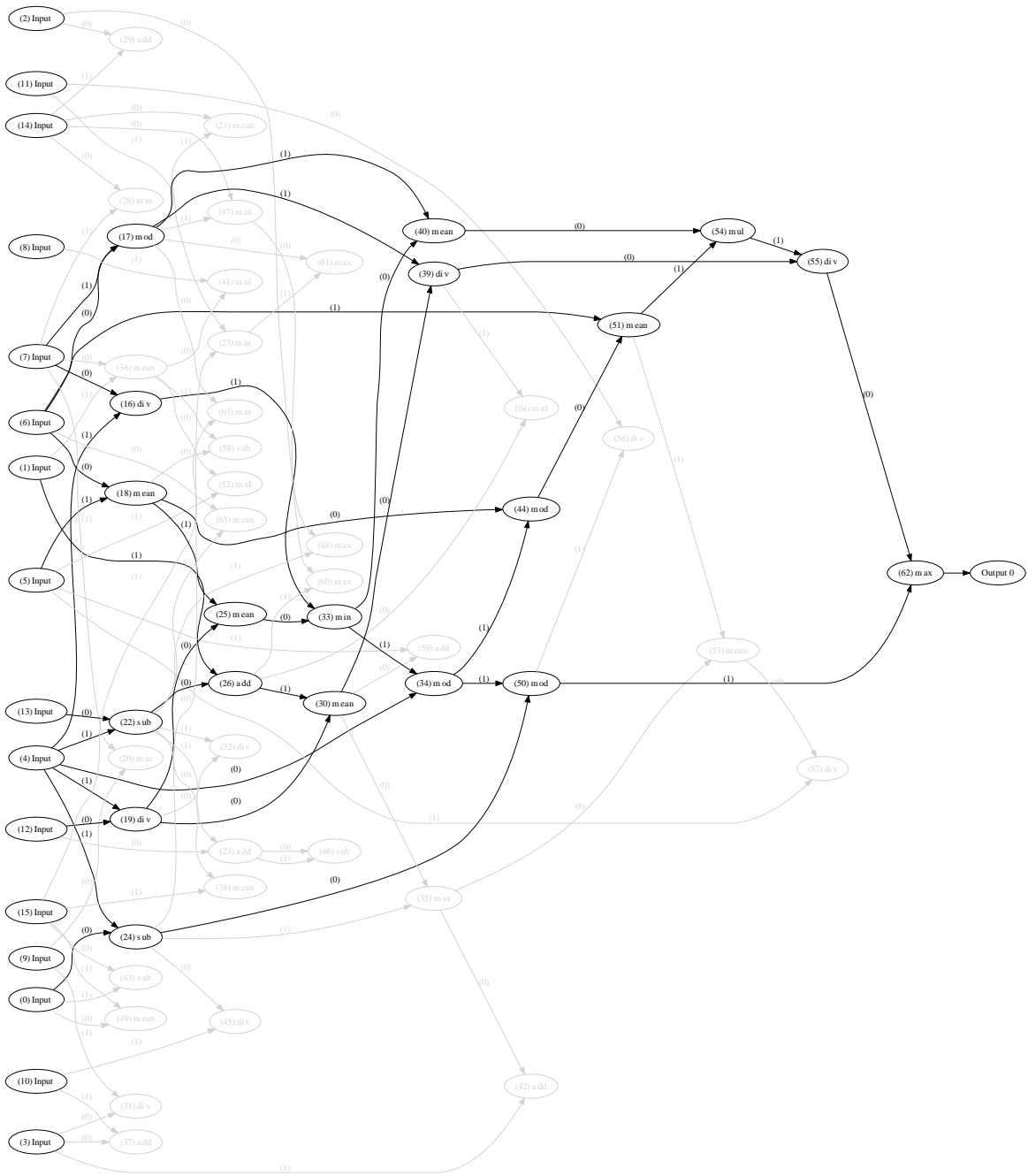

Figure 8 - CGP classification tree for the classification of prodromal PD vs. controls; example 2

PD Diagnosis

To evaluate the correspondence between participant group and DCM features, three mixed ANOVAs were conducted. In each ANOVA only two groups (e.g., PD patients and control) were considered rather than all three groups (PD patients, prodromal PD patients, and healthy control) at once, to mimic the classification method used in which, again, only two groups were considered at any point. For all three ANOVAs, a Greenhouse-Geisser correction was used as the models violated sphericity.

Firstly, to examine the correspondence between PD and control participants and DCM features, these were subjected to a mixed $2 \times 16$ ANOVA between the participant group (PD patients and healthy control) and DCM features (16 inputs per participant). The ANOVA revealed a significant main effect of DCM features $(F(5.47, 590.68) = 64.25, MSE = 2.17, p < .001, \eta_p^2 = .37)$. There was no significant interaction effect between participant group and DCM features $(F(5.47, 590.68) = 2.08, MSE = 0.07, p = .060, \eta_p^2 = .02)$ and no significant main effect of group $(F(1, 108) = 0.02, MSE = 0.00, p = .879, \eta_p^2 = .00)$.

Secondly, to explore the correspondence between PD patients and prodromal PD participants and DCM features, these were subjected to a mixed $2 \times 16$ ANOVA between the participant group (PD patients and prodromal PD) and DCM features (16 inputs per participant). The ANOVA revealed a significant main effect of DCM features $(F(5.66, 667.89) = 112.36, MSE = 3.68, p < .001, \eta_p^2 = .49)$. There was no significant interaction effect between participant group and DCM features $(F(5.66, 667.89) = 1.22, MSE = 0.04, p = .296, \eta_p^2 = .01)$ and no significant main effect of group $(F(1, 118) = 0.17, MSE = 0.00, p = .680, \eta_p^2 = .00)$.

Finally, to investigate the correspondence between prodromal PD and control participants and DCM features, these were subjected to a mixed $2 \times 16$ ANOVA between the participant group (prodromal PD and control) and DCM features (16 inputs per participant). The ANOVA revealed a significant main effect of DCM features $(F(5.15, 123.58) = 44.51, MSE = 1.71, p < .001, \eta_p^2 = .65)$. There was no significant interaction effect between participant group and DCM features $(F(5.15, 123.58) = 1.16, MSE = 0.05, p = .331, \eta_p^2 = .05)$ and no significant main effect of group $(F(1, 24) = 0.01, MSE = 0.00, p = .913, \eta_p^2 = .00)$.

Over the three ANOVAs, findings consistently revealed a main effect of DCM features with no significant main effect of participant group and no significant interaction effect. This represents the key finding as it indicates that the features in general are essential, whereas information on participant group *per se* is not.

### 3.3. *k*-Fold Cross-Validation

To evaluate the performance of the classifier, *k*-fold CV was conducted on all the different combinations of inputs for both DCM and timeseries values.

### 3.3.1. Cross-Validation for RCGP for Timeseries

The inputs were divided into folds with 80% of the data used for training, 10% for validation, and 10% for test. After the artificial data samples were synthesised for the minor class in the training set (using ADASYN), CV was repeated for 10 runs and the results were averaged, as depicted in Table 5, Table 6, and Table 7. Findings revealed that PD patients were successfully classified from healthy controls with a maximum accuracy of 91.22% using RCGP in CV (minimum accuracy: 87.55%, see Table 5). PD patients were successfully classified from prodromal PD patients with a maximum accuracy of 82.99% using RCGP in CV (minimum accuracy: 79.54%, see Table 6). Prodromal PD patients were successfully classified from healthy controls with a maximum accuracy of 68.28% using RCGP in CV (minimum accuracy: 62.58%, see Table 7).

PD Diagnosis

Table 5 - Cross-validation results for the timeseries values (PD vs. controls)

|  |  | Training % (*SD*) | Validation % (*SD*) | Test % (*SD*) |
|---|---|---|---|---|
| Classification results for each DMN region | | | | |
| PCC | RCGP | 88.95 (8.14) | 92.09 (0.45) | 88.42 (8.03) |
| mPFC | RCGP | 85.76 (8.49) | 92.60 (1.13) | 87.55 (6.40) |
| RIPC | RCGP | 91.45 (1.40) | 92.76 (1.48) | 90.99 (1.50) |
| LIPC | RCGP | 90.92 (1.70) | 92.02 (0.48) | 91.22 (0.51) |
| Classification results for all the DMN regions (4 inputs) | | | | |
|  | RCGP | 91.54 (0.86) | 92.49 (1.16) | 90.48 (2.16) |
| Classification results for all the DMN regions (1 input) | | | | |
|  | RCGP | 90.58 (2.61) | 92.37 (0.99) | 90.45 (1.62) |

Table 6 - Cross-validation results for the timeseries values (PD vs. prodromal PD)

|  |  | Training % (*SD*) | Validation % (*SD*) | Test % (*SD*) |
|---|---|---|---|---|
| Classification results for each DMN region | | | | |
| PCC | RCGP | 82.91 (2.46) | 85.20 (1.06) | 81.74 (3.18) |
| mPFC | RCGP | 82.44 (2.50) | 84.53 (0.28) | 82.56 (2.19) |
| RIPC | RCGP | 81.96 (4.15) | 84.97 (1.23) | 79.54 (7.56) |
| LIPC | RCGP | 82.81 (2.17) | 85.19 (0.89) | 82.59 (1.65) |
| Classification results for all the DMN regions (4 inputs) | | | | |
|  | RCGP | 81.82 (6.60) | 84.94 (0.85) | 80.96 (8.07) |
| Classification results for all the DMN regions (1 input) | | | | |
|  | RCGP | 82.68 (2.55) | 85.18 (1.11) | 82.99 (1.88) |

PD Diagnosis

Table 7 - Cross-validation results for the timeseries values (prodromal PD vs. controls)

|  |  | Training % (SD) | Validation % (SD) | Test % (SD) |
|---|---|---|---|---|
| Classification results for each DMN region | | | | |
| PCC | RCGP | 66.90 (4.33) | 75.93 (5.80) | 64.20 (7.01) |
| mPFC | RCGP | 70.91 (7.12) | 81.62 (6.70) | 67.15 (10.77) |
| RIPC | RCGP | 68.53 (7.74) | 77.25 (4.18) | 62.58 (11.13) |
| LIPC | RCGP | 65.87 (10.02) | 79.52 (5.36) | 68.28 (10.15) |
| Classification results for all the DMN regions (4 inputs) | | | | |
| | RCGP | 66.71 (5.64) | 79.78 (5.75) | 63.53 (9.79) |
| Classification results for all the DMN regions (1 input) | | | | |
| | RCGP | 63.49 (4.82) | 79.94 (6.26) | 64.64 (12.77) |

### 3.3.2. Cross-Validation for CGP for DCM

The inputs were divided into folds with 80% of the data used for training, 10% for validation, and 10% for test. After the artificial data samples were synthesised for the minor class in the training set (using ADASYN), CV was repeated for 10 runs and the results were averaged, as shown in Table 8.

Table 8 - Cross-validation results for DCM values

|  | Training % (SD) | Validation % (SD) | Test % (SD) |
|---|---|---|---|
| PD vs. controls | | | |
| CGP | 77.37 (10.28) | 85.63 (3.97) | 76.45 (8.43) |
| PD vs. prodromal PD | | | |
| CGP | 64.17 (9.56) | 75.38 (6.36) | 63.28 (6.01) |
| Prodromal PD vs. controls | | | |
| CGP | 59.91 (6.42) | 88.45 (9.58) | 53.81 (13.85) |

The results revealed that PD patients were successfully classified from the healthy controls with an accuracy of 76.45% using CGP in CV. PD patients were classified from the prodromal PD participants with 63.28% accuracy using CGP in CV. Finally, the findings indicated that prodromal participants were classified from healthy controls with 53.81% accuracy using CGP in CV.

## 4. Discussion

This research examined the classification of participants diagnosed with PD, prodromal PD participants, and healthy age-matched controls using rs-fMRI data. A key research question was: Can early stage PD (prodromal PD) be diagnosed using EAs on rs-fMRI data? Another distinctive element of this research involved (1) the application of EAs (CGP) as a classification method and (2) DCM analysis for classification. CGP classification was used for DCM analyses as well as timeseries analyses and the findings were validated with two other commonly used classification methods (ANN and SVM). A crucial benefit of EAs, specifically CGP, is that they provide a *white box solution* giving more information on the inputs used and enhanced knowledge concerning the final solution obtained in classification, compared to ANN and SVM. Findings were additionally validated using *k*-fold CV technique on the data.

The data was highly imbalanced, hence, ADASYN was used to balance the data before performing CV, resulting in an equal class distribution within the training set (i.e., balanced numbers of PD, prodromal PD, and control participants). Across timeseries and DCM analyses, findings revealed that PD versus control participants were classified with a maximum accuracy of 92%, PD versus prodromal PD participants with a maximum accuracy of 86%, and prodromal PD versus control participants with a maximum accuracy of 75%. These findings are notable as early diagnosis of PD (before motor symptoms) is in its infancy with high rates of misdiagnosis, impacting on patient treatment and quality of life. This finding embodies the most important research output from this research.

Findings further revealed almost no difference in the classification accuracies between timeseries and DCM data. In addition, findings revealed that CGP almost always provided equivalent performance accuracy when compared with ANN and SVM classification methods. Hence, these findings underscore the relevance of DCM analyses for classification and CGP as a novel classification tool for brain imaging data.

A novel question addressed by this research was: is DCM analysis useful for classification? Although DCM has recently become a widespread tool to model effective connectivity in neuroimaging data [34], [54]–[57], no research has examined classification using DCM analysis for any type of fMRI data, including rs-fMRI. Moreover, few studies have conducted

PD Diagnosis

DCM analysis on PD data [30]–[35]. Hence, this research aimed to further existing knowledge on DCM as applied to PD. An interesting finding was little difference in the classification accuracies between timeseries and DCM analyses (1-6% difference in the maximum accuracies). The findings, firstly, speak to the relevance of DCM data for classification and, secondly, broaden the small literature on DCM as applied to PD data. DCM analysis was conducted on the DMN, yet, new research [58] has examined DCM across multiple brain networks and currently whole-brain DCM of fMRI data is an ongoing project by the SPM developers lab. As such, exploring DCM across specific regions influenced by PD (e.g., basal nigra) is an exciting avenue for future research.

Classification of rs-fMRI data is a key theme examined by this research, specifically, this research explored the applicability of CGP and RCGP to classify rs-fMRI data. Previous research on the classification of fMRI data has largely focused on statistical modelling techniques (e.g., independent components analysis, multivariate pattern analysis; [59]–[61]). These latter approaches are mostly predictive (hypothesis driven), yet, machine learning methods (as per this research) are explanatory modelling techniques (mostly data driven). Machine learning approaches are advantageous given that (1) they can be validated, providing an unbiased estimate of accuracy [62]; (2) they are typically based on fewer assumptions (relative to statistical-based techniques); and (3) they can predict and learn concurrently from large datasets, whereas statistical modelling are typically used for small datasets to avoid overfitting [59], [60].

The approach used in this research was distinctive as EAs, specifically CGP, have not been used for classification of brain imaging data. CGP provided approximately equivalent performance accuracy when compared with ANN and SVM classification methods across all participant groups (PD, prodromal PD, and control). Further research can examine CGP as applied to task-based fMRI, which would enable researchers to explore a number of sophisticated questions including finger-tapping tasks (unlike rs-fMRI), which is used in behavioural measures and is currently a leading method for confirming PD diagnosis.

Following *k*-fold CV, classification accuracy for timeseries values decreased by only a few percentage. Yet, for DCM values, accuracy was reduced to 76% for the PD versus control participant group, although these findings held across all participant groups. The reduced

PD Diagnosis

accuracy is due to having used class-imbalanced data as, whilst there was a large sample of PD patients, sample sizes of the prodromal PD and control groups were comparably smaller. Moreover, the number of features included in the classification was significantly smaller in DCM values in comparison with the timeseries values.

Imbalanced data limits classification accuracy [36], [63], [64] as most classification methods assume balanced class distributions, creating two problems. Firstly, high class-imbalanced data results in limited training set sample size as classifiers often treat imbalanced data as though they have small sample sizes (since some classes may have small numbers). Secondly, learning focuses on classes with larger sample sizes, rather than focusing on discriminating between the sizes of the classes in the data and the characteristics of the actual data (rather than the synthesised data). Solutions to imbalanced data typically include changing/generating data to obtain equal class-distributions, which tends to improve classification accuracy [38]–[40]. Nonetheless, given that the synthetic balanced data is applied only to the training sets, classification accuracy is limited for the validation and test sets as before the classifier that was trained for the CV was a different classifier with a different accuracy level than that used in the classification part. This is an inevitable problem associated with imbalanced data, regardless of the number of folds involved in CV.

In this research, ADASYN was used, which is a method of generating synthetic data to create balanced class-distributions for the training set (i.e., generating balanced numbers of PD, prodromal PD, and control participants). ADASYN was applied to the three classification groups: PD versus control, PD versus prodromal PD, and prodromal PD versus control. Findings revealed maximum classification accuracies of 71-92% across all groups when using timeseries and DCM data. Despite the reduced accuracy following CV, results revealed that classification was reliable for both timeseries and DCM values. Hence, classification on DCM and timeseries values can be used as brain-imaging biomarkers for PD and the current findings underscore the relevance of CGP as a classification method, even for highly imbalanced data.

This study involved an analysis of a large open dataset (PPMI) [25]. Contributing to open science (e.g., making data publically available) shares resources from one project to other research, such as the re-analysis of biomedical raw data to examine new predictions [65], as per the current research. Advantages of open science include data sharing, saving resources (e.g., money and time), exchanging expert viewpoints, reducing fraud and *p*-hacking, training purposes, amongst others. Here, in addition to the reproducibility and resource-sharing advantages provided by open data, this research was able to access a sample of 128 participants (excluding participants who did not fulfil the eligibility requirements). Most previous research using fMRI data acquire limited samples, typically 20 participants ([45], [66]–[68]; although 77 participants were tested in [46]). Having such a large sample size lends confidence to the generalisability of these findings.

## 5. Conclusion

To conclude, this research explored the classification of rs-fMRI data in participants diagnosed with PD, prodromal PD participants, and healthy age-matched controls using novel data (DCM analyses) and a novel classification method (CGP). Classification accuracy for DCM analyses was compared to that of timeseries analyses, and two other classification methods (ANN and SVM) were used to validate the findings, as well as employing *k*-fold CV. Across DCM and timeseries analyses, findings revealed maximum accuracies of 86% for classification of prodromal PD patients versus PD patients, and 92% for PD patients versus healthy controls using CGP. Early diagnosis of PD (before motor symptoms appear) is rife with challenges and current methods have high rates of misdiagnosis. This research further revealed a maximum accuracy of 75% in differentiating prodromal PD patients from healthy controls, with medical implications for disease management, patient treatment, and patient quality of life. Hence, this finding is the most important research output from this research. Furthermore, classification accuracy was approximately equivalent for (1) DCM analyses and timeseries analyses and (2) different classification methods (CGP, ANN, or SVM). CGP has distinct advantages regarding the information linked to the solutions they generate. Therefore, these findings speak to the relevance of CGP as a novel classification tool for two types of brain imaging data (DCM and timeseries rs-fMRI analyses). This research developed automatic procedures for identifying PD brain imaging preclinical biomarkers, which are fundamental in improving accuracy of PD diagnosis methods. Furthermore, these findings highlight the applicability of DCM analyses for classification.

PD Diagnosis[10] L. Findley *et al.*, "Direct economic impact of Parkinson's disease: A research survey in the United Kingdom," *Mov. Disord.*, vol. 18, no. 10, pp. 1139–1145, 2003.

[11] K. Whetten-Goldstein, F. Sloan, E. Kulas, T. Cutson, and M. Schenkman, "The burden of Parkinson's disease on society, family, and the individual.," *Journal of the American Geriatrics Society*, vol. 45, no. 7. pp. 844–9, 1997.

[12] C. LePen, S. Wait, F. Moutard-Martin, M. Dujardin, and M. Ziégler, "Cost of illness and disease severity in a cohort of French patients with Parkinson's disease," *Pharmacoeconomics*, vol. 16, no. 1, pp. 59–69, 1999.

[13] R. C. Dodel *et al.*, "Cost of illness in Parkinson disease. A retrospective 3-month analysis of direct costs," *Nervenarzt*, vol. 68, no. 12, pp. 978–984, Dec. 1997.

[14] C. G. Goetz *et al.*, "Movement Disorder Society-sponsored revision of the Unified Parkinson's Disease Rating Scale (MDS-UPDRS): Process, format, and clinimetric testing plan.," *Mov. Disord.*, vol. 22, no. 1, pp. 41–7, Jan. 2007.

[15] Royal College of Physicians and Association of British Neurologists, *Local adult neurology services for the next decade*, no. June. London: Royal College of Physicians, 2011.

[16] C. B. Levine, K. R. Fahrbach, A. D. Siderowf, R. P. Estok, V. M. Ludensky, and S. D. Ross, "Diagnosis and Treatment of Parkinson's Disease: A Systematic Review of the Literature," *AHRQ*, vol. 03, no. 57, p. 306, 2003.

[17] A. J. Hughes, S. E. Daniel, and A. J. Lees, "Improved accuracy of clinical diagnosis of Lewy body Parkinson's disease," *Neurology*, vol. 57, no. 8, pp. 1497–1499, Oct. 2001.

[18] A. J. Hughes, S. E. Daniel, Y. Ben-Shlomo, and A. J. Lees, "The accuracy of diagnosis of parkinsonian syndromes in a specialist movement disorder service.," *Brain a J. Neurol.*, vol. 125, no. Pt 4, pp. 861–870, 2002.

[19] J. Jankovic, A. H. Rajput, M. P. McDermott, and D. P. Perl, "The evolution of diagnosis